%% file: acl_latex.tex
\pdfoutput=1

\documentclass[11pt]{article}

\usepackage[final]{acl}

\usepackage{times}
\usepackage{latexsym}

\usepackage[T1]{fontenc}

\usepackage[utf8]{inputenc}

\usepackage{microtype}

\usepackage{inconsolata}

\usepackage{graphicx}

\usepackage{booktabs}
\usepackage{multirow}
\usepackage{amsmath}
\usepackage{amsfonts}
\usepackage{kotex}
\usepackage{xcolor}
\usepackage{algorithm}
\usepackage{algorithmic}
\usepackage{tcolorbox}
\usepackage{fancyvrb}
\tcbuselibrary{skins,breakable}
\usepackage{listings}
\usepackage{fvextra}

\title{EdiText: Controllable Coarse-to-Fine Text Editing with \mbox{Diffusion Language Models}}

\author{{\bf Che Hyun Lee$^{1}$} \hspace{6mm}  {\bf Heeseung Kim$^{1}$} \hspace{6mm}  {\bf Jiheum Yeom$^{1}$} \hspace{6mm}  {\bf Sungroh Yoon$^{1,2}$\Thanks{\hspace{0.1em} Corresponding Author}} \\ \\
   $^{1}$Department of Electrical and Computer Engineering, Seoul National University \\
   $^{2}$AIIS, ASRI, INMC, ISRC, and IPAI, Seoul National University \\ \\
   {\tt \fontsize{10}{10}\selectfont \{saga1214, gmltmd789, quilava1234, sryoon\}@snu.ac.kr}}

\begin{document}
\maketitle
\begin{abstract}
We propose EdiText, a controllable text editing method that modifies the reference text to desired attributes at various scales. We integrate an SDEdit-based editing technique that allows for broad adjustments in the degree of text editing. Additionally, we introduce a novel fine-level editing method based on self-conditioning, which allows subtle control of reference text. While being capable of editing on its own, this fine-grained method, integrated with the SDEdit approach, enables EdiText to make precise adjustments within the desired range. EdiText demonstrates its controllability to robustly adjust reference text at a broad range of levels across various tasks, including toxicity control and sentiment control. 

\end{abstract}

\input{sections/introduction}

\input{sections/related_works}

\input{sections/methods}

\input{sections/experiments}

\input{sections/results}

\input{sections/conclusion}

\input{sections/acknowledgement}

\input{sections/limitations}

\bibliography{acl_latex}

\appendix
\input{sections/appendix}

\end{document}

%% file: sections/introduction.tex
\section{Introduction}

Editing tasks, which involve modifying given reference data in a desired direction, have been widely explored \cite{bar2022text2live, 9257074}, providing versatility and convenience. With recent advances in diffusion models \cite{DDPM} across various domains \cite{videoworldsimulators2024, evans2024fast, Rombach_2022_CVPR}, numerous diffusion-based editing methods have been introduced for various types of data with diverse attributes \cite{hertz2022prompt, NEURIPS2023_e1b619a9}. Specifically, diverse approaches have been proposed for editing visual data at various levels, from global attributes \cite{meng2022sdedit} to fine-grained details \cite{kawar2023imagic, mokady2022null}.

Editing methods have also emerged in the text domain \cite{du-etal-2022-read, mallinson-etal-2020-felix, malmi-etal-2019-encode, Martinez_Dass_Kurohashi_Jurafsky_Yang_2020}.
Although autoregressive (AR) language models \cite{NEURIPS2020_1457c0d6, JMLR:v21:20-074, touvron2023llama} generally outperform non-autoregressive (NAR) counterparts \cite{Clark2020ELECTRA:, devlin-etal-2019-bert, li2022diffusionlm, liu2019roberta}, NAR models have shown strengths in controlled generation \cite{li2022diffusionlm, mireshghallah-etal-2022-mix}, where shifting to the target distribution can be easily achieved with a relatively small amount of data and training.
The ability of NAR models to facilitate such shifts can also be effectively applied to text editing.  For example, one might need an edit to shift the overall tone of a report from informal to formal, or perform a local edit to change only specific words. In line with these possibilities, a number of studies \cite{Horvitz_Patel_Callison-Burch_Yu_McKeown_2024, mallinson-etal-2020-felix, mireshghallah-etal-2022-mix, Martinez_Dass_Kurohashi_Jurafsky_Yang_2020} have adapted NAR-based controlled generation methods to text editing. 
Although these models are capable of reflecting the target attributes in text to some extent, their editing techniques either lack the ability to control the degree of editing or are limited to a narrow range, which restricts their versatility.

\begin{figure*}
    \centering
    \includegraphics[width=1.0\linewidth]{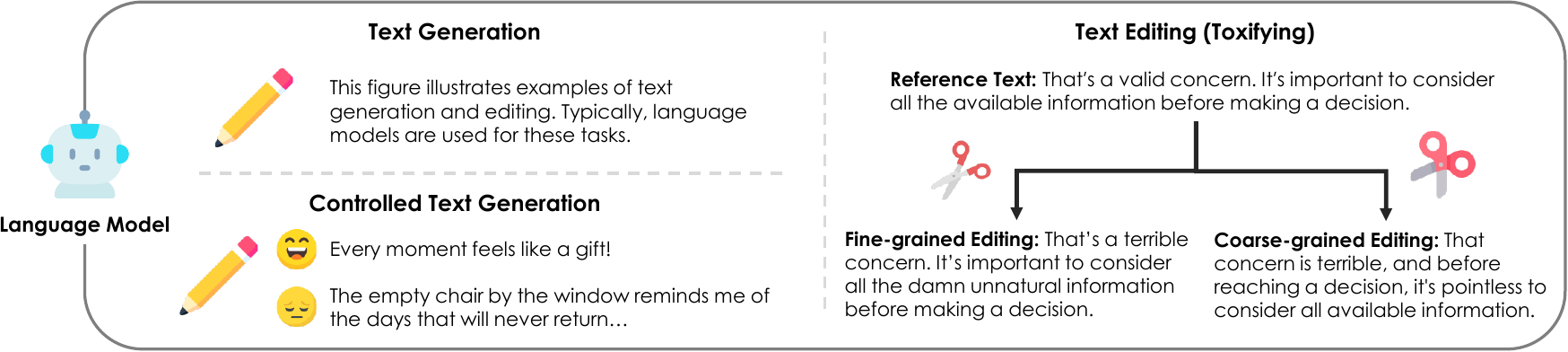}
    \caption{Comparisons of various text-based tasks requiring control. Our text editing method focuses on adjusting degree of editing, enabling both coarse- and fine-grained modification.}
    \label{fig:TextEditing}
    \vskip -0.2in
\end{figure*}

Recently, the advancement of diffusion models has extended into the discrete text domain, which can be categorized into two main types \cite{zou2023survey}: discrete diffusion \cite{austin2021structured, he-etal-2023-diffusionbert, hoogeboom2021argmax, NEURIPS2024_eb0b13cc, NEURIPS2024_bad233b9} and embedding diffusion \cite{gao2022difformer, gong2022diffuseq, li2022diffusionlm}. Embedding diffusion models, specifically, map discrete tokens to continuous embeddings and model them through a Gaussian diffusion process, able to leverage various advancements from the diffusion models in other continuous domains \cite{li2022diffusionlm, lovelace2023latent, zhang2024language}. 
While this advantage demonstrates the potential of applying several techniques from continuous diffusion models \cite{chen2023analog, dhariwal2021diffusion, ho2021classifierfree}, diffusion-based text editing has not yet been sufficiently explored.

In this work, we propose EdiText, an embedding diffusion-based text editor that supports both coarse- and fine-grained editing. We adapt SDEdit \cite{meng2022sdedit}, originally for images, to enable global-level text modifications. Furthermore, by reinterpreting self-conditioning \cite{chen2023analog}, a technique to enhance the performance of embedding diffusion, we introduce a novel fine-grained text editing approach that allows precise adjustments in the degree of editing. Additionally, by integrating the proposed coarse-grained and fine-grained editing techniques, our approach offers more comprehensive coverage. This integration bridges the gaps within the broad coverage of coarse-level editing by enabling fine-level adjustments, ensuring a more complete and precise editing process.

We demonstrate the effectiveness of our proposed editing methods across various tasks, achieving superior editing performance compared to the baseline model. We additionally showcase our model's capability to edit across a wider and more nuanced range by combining both techniques.

%% file: sections/related_works.tex
\section{Related Works}

\textbf{Diffusion Language Model.} Diffusion models have achieved remarkable performance in vision and audio tasks—especially in generation and editing. Building on these successes, there has been a growing interest in extending the diffusion framework to the text domain. A number of studies \cite{austin2021structured, chen2023analog, hoogeboom2021argmax, pmlr-v235-lou24a, lovelace2023latent, NEURIPS2024_eb0b13cc, NEURIPS2024_bad233b9} have adopted diffusion framework for text generation and confirmed their effectiveness. Moreover, several works \cite{gong2022diffuseq, li2022diffusionlm, zhang2024language} have highlighted the controllability of these models in constrained scenarios-such as tasks involving parse trees or fixed sequence lengths-by leveraging various guidance techniques \cite{dhariwal2021diffusion, ho2021classifierfree} tailored for diffusion frameworks. Despite these strengths, the potential of diffusion language models for text editing—a task that could benefit significantly from their broad controllability—remains underexplored.

\textbf{Controlled Text Generation.} Many studies have focused on methods to control the generation process to align with user intentions. One prominent task, controlled text generation, seeks to produce text that adheres to specific constraints (i.e. fixed sequence length, parse-tree, or specific target attributes).
While some approaches leverage AR models for controlled text generation \cite{carlsson-etal-2022-fine, yang-etal-2023-tailor}, NAR models \cite{li2022diffusionlm, mireshghallah-etal-2022-mix, zhang-etal-2024-languageflow} often require fewer parameters and benefit from a parallel sampling process. This parallelism enables NAR models to consider the entire context of the textual input, facilitating smoother distribution shifts during the sampling process.

\textbf{Text Editing.} A task similar to controlled text generation is text editing, which involves modifying a given reference text to align with the user's intent. Text editing differs from controlled text generation in that it focuses on transforming existing content based on specific goals, rather than generating entirely new text from scratch. Several studies have proposed reference-based editing methods \cite{du-etal-2022-read, malmi-etal-2019-encode, laban2023chat, raheja-etal-2023-coedit}, while some are also known as controlled paraphrase generation \cite{Dehghani2021ControllablePG, ogasa-etal-2024-controllable, yang-etal-2022-gcpg} which focuses on specific constraints. However, these approaches generally lack mechanisms to control the degree of editing, making it difficult to achieve the desired level of transformation.

\begin{figure*}
    \centering
    \includegraphics[width=1.0\linewidth]{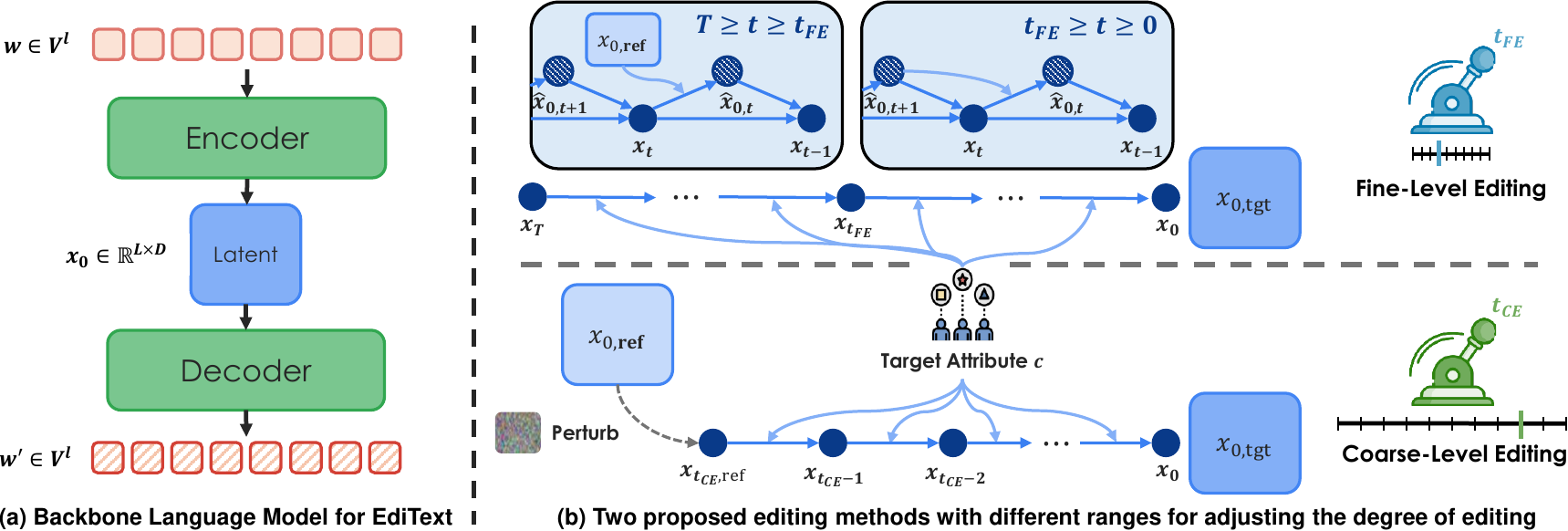}
    \caption{Overview of the proposed EdiText framework. (a) We employ an embedding diffusion model as the backbone language model for EdiText (Section \ref{subsection:latent diffusion}). (b) The two editing methods with different degree of editing—Coarse-Level Editing and Fine-Level Editing—are described in Sections \ref{subsection:global editing} and \ref{subsection:local editing}, respectively.}
    \label{fig:overview}
    \vskip -0.2in
\end{figure*}

Some recent works have addressed this limitation. For example, \citet{mireshghallah-etal-2022-mix} employed an energy-based model (EBM) to adjust the degree of editing by weighting multiple target attributes during generation. Similarly, ParaGuide \cite{Horvitz_Patel_Callison-Burch_Yu_McKeown_2024} is a diffusion-based framework that uses classifier guidance to modulate editing intensity. While these methods can shift text toward the desired attribute, their control remains limited to micro-level adjustments. In contrast, our proposed model, EdiText, integrates diverse editing strategies within the diffusion framework, enabling text editing across multiple scales. The distinctions between general text generation, controlled text generation, and the varying-scale text editing targeted by our model are summarized in Figure \ref{fig:TextEditing}.

%% file: sections/methods.tex
\section{EdiText}

In this section, we introduce EdiText, a text editing framework which utilizes a diffusion model capable of performing both fine-grained and coarse-grained editing. We adopt Latent Diffusion for Language Generation (LD4LG) \cite{lovelace2023latent}, an embedding diffusion model, as the backbone of our framework to support text editing where the discrete text is transformed into continuous data, allowing diffusion-based editing methods, as described in Section \ref{subsection:latent diffusion}.
Inspired by the image editing technique SDEdit \cite{meng2022sdedit}, we present a method for controlling text editing at a coarse level in Section \ref{subsection:global editing}. 
Furthermore, to support fine-grained control of the extent of editing, we propose a novel self-conditioning-based approach \cite{chen2023analog} in Section \ref{subsection:local editing}. 
By combining these elements, EdiText is capable of editing text of varying attributes at different levels of granularity.
The overview of our proposed editing framework, EdiText, is illustrated in Figure \ref{fig:overview}.

\subsection{Latent Diffusion for Language Generation}
\label{subsection:latent diffusion}

Embedding diffusion model is a type of diffusion model for modeling discrete data. The model consists of two main components: the embedding module, which maps discrete tokens to continuous representations, and the diffusion process, which models the distribution of these representations. Latent Diffusion for Language Generation (LD4LG) \cite{lovelace2023latent} uses a language autoencoder as the embedding module, composed of a compression network $E$ to convert a discrete token sequence $w$ into latent representations $x$, and a decoder $D$ to reconstruct $w$.

The language encoder $E$, built with the Perceiver Resampler \cite{alayrac2022flamingo} architecture, compresses variable-length text tokens into a fixed-length continuous latent representation. More specifically, given an input token sequence $w \in V^l$ with the set of vocabulary $V$ and length $l$, the encoder converts the sequence into a fixed size latent representation $x \in \mathbb{R}^{L\times D}$ with length $L < l$ and hidden size $D$. These representations can then be reconstructed into variable-length sequences through an autoregressive decoder $D$. This feature enables fixed-size latent representations to generate text sequences of varying lengths, making it particularly suitable for text editing scenarios that require flexible control over differing input and output lengths. Moreover, by representing discrete text data as continuous representations, editing methods from the continuous domain—such as SDEdit \cite{meng2022sdedit}—can be effectively applied to the text domain. Given these advantages, we adopt LD4LG as the backbone for our editing framework.

To model the distribution of the latent representation, \citet{lovelace2023latent} first define a forward diffusion process which gradually transforms $x_0=x$ into random noise $x_T\sim N(0,I)$ over timesteps $T$. This forward process is defined by a noise schedule $\alpha_t$, and is utilized to compute the corrupted representation $x_t=\sqrt{\alpha_t}x_0 + \sqrt{1-\alpha_t}\epsilon_t$ at an intermediate timestep $t\in[1,T]$, where $\epsilon_t$ is a noise vector that follows a standard normal distribution. To sample the latent representation $x_0$, \citet{lovelace2023latent} follow the reverse trajectory of the pre-defined forward process learned by a diffusion model. 
Unlike conventional diffusion approaches that model the noise vector $\epsilon_t$ and reconstruct data from noisy samples, they employ a reparametrization strategy: they train a network $x_\theta$ to predict the clean latent $x_0$ directly from noisy input $x_t$. 

The training objective for LD4LG minimizes the following loss:

\vskip -0.2in
\begin{align}   
    \label{equation:loss}
        L(\theta)&={\mathbb{E}_{t,x_0,\epsilon_t}[\lambda_t^{-1}\lVert x_\theta(x_t,t) - x_0\rVert_2^2]}.
\end{align}

\noindent where $\lambda_t=1-\alpha_t$.

At inference time, LD4LG iteratively denoise according to

\vskip -0.2in
\begin{align}   
        \label{equation:sampling}
        x_{t-1} &= \sqrt{\alpha_{t-1}} x_\theta(x_t,t) + \sqrt{\lambda_{t-1}}\epsilon_\theta(x_t,t), \\
        \label{equation:conversion}
        \epsilon_\theta(x_t,t) &= \lambda_t^{-0.5}(x_t-\sqrt{\alpha_t}x_\theta(x_t,t)).
\end{align}

\noindent where $\epsilon_\theta(x_t,t)$ is the reparameterized version of $x_\theta(x_t,t)$. 

\begin{algorithm}[tb]
\caption{Our Coarse-Level Editing}
\label{alg:global editing}
\textbf{Input:} diffusion model $x_\theta(x_t,t,\cdot)$, noise schedule $\alpha_t$, language encoder and decoder $E(\cdot)$ and $D(\cdot)$, total timestep $T$, reference text $w_{ref}$, timestep for coarse-level editing $t_{CE}$, target attribute $c$

\textbf{Output:} edited text \(w_{e}\)

\begin{algorithmic}[1] 
\STATE $x_0 = E(w_{ref})$ \quad\# encoded reference text
\STATE $\lambda_{t_{CE}}=1-\alpha_{t_{CE}}$
\STATE $\epsilon_{t_{CE}} \sim N(0,I)$
\STATE $x_{t_{CE}} = \sqrt{\alpha_{t_{CE}}}x_0 + \sqrt{\lambda_{t_{CE}}}\epsilon_{t_{CE}}$
\FOR{$t = t_{CE}, ..., 1$}
\IF{$t = t_{CE}$}
\STATE $\hat{x}_{0,t}=x_\theta(x_t,t,\emptyset, c)$
\ELSE
\STATE $\hat{x}_{0,t}=\hat{x}_{0,t+1,SC}$ \quad\# SC for self-cond
\ENDIF
\STATE $\lambda_t=1-\alpha_t$
\STATE $\hat{x}_{0,t,SC}=x_\theta(x_t,t,\hat{x}_{0,t}, c)$
\STATE $\epsilon_\theta(x_t,t,\hat{x}_{0,t}, c)=\lambda_t^{-0.5}(x_t-\sqrt{\alpha_t}\hat{x}_{0,t,SC})$
\STATE $x_{t-1} = \sqrt{\alpha_{t-1}} \hat{x}_{0,t,SC} + \sqrt{\lambda_{t-1}}\epsilon_\theta(x_t,t,\hat{x}_{0,t}, c)$
\ENDFOR
\STATE \textbf{return} $w_e=D(x_0)$
\end{algorithmic}
\end{algorithm}

To improve sample quality, LD4LG leverages the self-conditioning \cite{chen2023analog}, which uses the previous timestep's prediction as an additional condition for the diffusion model at current timestep. During training, the diffusion model alternates between two distinct modes with probability $p$. Specifically, at timestep $t$, the diffusion model $x_\theta(x_t,t,\emptyset)$ is trained using Eq. \ref{equation:loss} with a pre-defined probability $p$. Here, $\emptyset$ indicates that, as in a typical diffusion model, no extra condition is provided beyond the current timestep $t$ and the corrupted data $x_t$ when predicting $x_0$. Conversely, with probability $1-p$, the model employs the prediction $\hat{x}_{0,t}=x_\theta(x_t,t,\emptyset)$ as a condition, thereby training $x_\theta(x_t,t,\hat{x}_{0,t})$ to predict $x_0$. During self-conditioning-based training, gradients do not propagate back through the condition $\hat{x}_{0,t}$, and the loss function is defined as follows:
\begin{align}   
    \label{equation:loss for self-conditioning}
        L={\mathbb{E}_{t,x_0,\epsilon_t}[(1-\alpha_t)^{-1}\lVert x_\theta(x_t,t,\hat{x}_{0,t}) - x_0\rVert_2^2]}.
\end{align}

\begin{algorithm}[tb]
\caption{Our Fine-Level Editing}
\label{alg:local editing}
\textbf{Input:} diffusion model $x_\theta(x_t,t,\cdot)$, noise schedule $\alpha_t$, language encoder and decoder $E(\cdot)$ and $D(\cdot)$, total timestep $T$, reference text $w_{ref}$, timestep for fine-level editing $t_{FE}$, target attribute $c$

\textbf{Output:} edited text \(w_{e}\)
\begin{algorithmic}[1] 
\STATE $x_T = z \sim N(0,I)$
\FOR{$t = T, ..., 1$}    
\IF{$t_{FE} \leq t \leq T$}
\STATE $\hat{x}_{0,t}=E(w_{ref})$
\ELSE
\STATE $\hat{x}_{0,t}=\hat{x}_{0,t+1,SC}$
\ENDIF
\STATE $\lambda_t=1-\alpha_t$
\STATE $\hat{x}_{0,t,SC}=x_\theta(x_t,t,\hat{x}_{0,t}, c)$
\STATE $\epsilon_\theta(x_t,t,\hat{x}_{0,t}, c)=\lambda_t^{-0.5}(x_t-\sqrt{\alpha_t}\hat{x}_{0,t,SC})$
\STATE $x_{t-1} = \sqrt{\alpha_{t-1}} \hat{x}_{0,t,SC} + \sqrt{\lambda_{t-1}}\epsilon_\theta(x_t,t,\hat{x}_{0,t}, c)$
\ENDFOR
\STATE \textbf{return} $w_e=D(x_0)$
\end{algorithmic}
\end{algorithm}

During inference, LD4LG generates improved-quality latent representations with self-conditioning by recursively using the prediction from the previous timestep $t+1$ for denoising. At timestep $T$, since  no previous timestep exists, sampling is performed without self-conditioning. We set $p=0.5$ following \citet{chen2023analog}.

\begin{table*}[ht]
    \fontsize{8}{11}\selectfont
    \centering
    \begin{tabular}{l|ccc|cc}
        \toprule
        \multirow{2}{*}{\textbf{Method}}    & \multicolumn{3}{c|}{\textbf{Retention Rate}} & \multicolumn{2}{c}{\textbf{Toxicity}} \\
        \cmidrule(lr){2-4} \cmidrule(lr){5-6}
                                   & \textbf{Hamming} $(\downarrow)$        & \textbf{SacreBLEU} $(\uparrow)$  & \textbf{BERTScore} $(\uparrow)$  & \textbf{Moderation} $(\downarrow)$ & \textbf{PerspectiveAI} $(\downarrow)$ \\
        \midrule
        Reference (toxic)           & 0.000               & 100.000                & 1.000               & 0.810      & 0.708    \\
        \midrule
        ParaGuide ($\lambda=200$)                   & 25.303          & 14.935           & 0.903            & 0.446      & 0.321    \\
        ParaGuide ($\lambda=1,000$)                  & 26.221          & 12.791           & 0.896            & 0.413      & 0.281    \\
        ParaGuide ($\lambda=10,000$)                 & 27.166          & 10.968           & 0.889            & 0.335      & 0.229    \\
        \midrule
        Qwen2.5-0.5B-Instruct                       & 27.193          & 31.119           & 0.903            & 0.347      & 0.312    \\
        Qwen2.5-0.5B-Instruct (Amp.)                 & 26.873          & 30.817           & 0.907            & 0.359      & 0.331    \\
        \midrule
        EdiText-CE ($t_{CE}=125$)                   & 6.273           & 72.922           & 0.977            & 0.813      & 0.601    \\
        EdiText-CE ($t_{CE}=150$)                   & 10.111          & 58.799           & 0.960            & 0.790      & 0.579    \\
        EdiText-CE ($t_{CE}=175$)                   & 17.412          & 34.722           & 0.923            & 0.576      & 0.450    \\
        EdiText-CE ($t_{CE}=200$)                   & 28.906          & 7.640            & 0.865            & 0.105      & 0.136    \\
        EdiText-CE ($t_{CE}=225$)                   & 34.138          & 0.853            & 0.847            & 0.003      & 0.046    \\
        \midrule
        EdiText-FE ($t_{FE}=25$)                    & 24.745          & 14.868           & 0.881            & 0.117      & 0.121    \\
        EdiText-FE ($t_{FE}=75$)                    & 25.531          & 13.406           & 0.878            & 0.112      & 0.115    \\
        EdiText-FE ($t_{FE}=125$)                   & 27.832          & 9.572            & 0.869            & 0.073      & 0.094    \\
        EdiText-FE ($t_{FE}=175$)                   & 31.846          & 4.168            & 0.857            & 0.024      & 0.059    \\
        \bottomrule
    \end{tabular}
    \vskip -0.1in
    \caption{Quantitative results for coarse- and fine-level detoxifying tasks on toxic data. \textbf{Hamming} represents the Hamming Distance. $\lambda$ denotes the guidance strength for ParaGuide, and Amp. refers to the amplified prompt setup.}
    \label{tab:table1}
    \vskip -0.2in
\end{table*}

To better adapt LD4LG for text editing, we train a class-conditional LD4LG that accepts a desired attribute as condition. Given a pre-defined class label $c$, i.e. toxifying or detoxifying label in a toxicity task, we replace $x_\theta(x_t,t,\emptyset)$ and $x_\theta(x_t,t,\hat{x}_{0,t})$ with $x_\theta(x_t,t,\emptyset,c)$ and $x_\theta(x_t,t,\hat{x}_{0,t}, c)$ during training. This change lets LD4LG generate text with the target condition, thereby extending its application to text editing by incorporating our proposed editing methods, as described in the following section.

\subsection{SDEdit-Based Coarse-Level Editing}
\label{subsection:global editing}

\begin{table*}[ht]
    \fontsize{8}{11}\selectfont
    \centering
    \begin{tabular}{l|ccc|cc}
        \toprule
        \multirow{2}{*}{\textbf{Method}}    & \multicolumn{3}{c|}{\textbf{Retention Rate}} & \multicolumn{2}{c}{\textbf{Toxicity}} \\
        \cmidrule(lr){2-4} \cmidrule(lr){5-6}
                                   & \textbf{Hamming} $(\downarrow)$        & \textbf{SacreBLEU} $(\uparrow)$  & \textbf{BERTScore} $(\uparrow)$  & \textbf{Moderation} $(\uparrow)$ & \textbf{PerspectiveAI} $(\uparrow)$ \\
        \midrule
        Reference (Nontoxic)             & 0.000               & 100.000                & 1.000               & 0.001      & 0.000    \\
        \midrule
        ParaGuide ($\lambda=200$)        & 25.612          & 18.526           & 0.919            & 0.073      & 0.151    \\
        ParaGuide ($\lambda=1,000$)       & 29.470          & 10.234           & 0.896            & 0.163      & 0.267    \\
        ParaGuide ($\lambda=10,000$)      & 34.960          & 2.746            & 0.866            & 0.285      & 0.367    \\
        \midrule
        Qwen2.5-0.5B-Instruct            & 29.904          & 32.121           & 0.920            & 0.002      & 0.040    \\
        Qwen2.5-0.5B-Instruct (Amp.)      & 31.829          & 29.440           & 0.909            & 0.000      & 0.037    \\
        \midrule
        EdiText-CE ($t_{CE}=125$)        & 7.247           & 71.251            & 0.978            & 0.005      & 0.041     \\
        EdiText-CE ($t_{CE}=150$)        & 10.594          & 58.056            & 0.964            & 0.015      & 0.052     \\
        EdiText-CE ($t_{CE}=175$)        & 17.236          & 37.455            & 0.934            & 0.103      & 0.113     \\
        EdiText-CE ($t_{CE}=200$)        & 29.257          & 10.030            & 0.876            & 0.553      & 0.385     \\
        EdiText-CE ($t_{CE}=225$)        & 36.135          & 0.792             & 0.848            & 0.851      & 0.604     \\
        \midrule
        EdiText-FE ($t_{FE}=25$)         & 24.894          & 18.306            & 0.894            & 0.474      & 0.375     \\
        EdiText-FE ($t_{FE}=75$)         & 25.550          & 16.584            & 0.890            & 0.494      & 0.388     \\
        EdiText-FE ($t_{FE}=125$)        & 27.817          & 11.724            & 0.878            & 0.607      & 0.445     \\
        EdiText-FE ($t_{FE}=175$)        & 32.331          & 5.167             & 0.860            & 0.782      & 0.562     \\
        \bottomrule
    \end{tabular}
    \vskip -0.1in
    \caption{Quantitative results for coarse- and fine-level toxifying tasks on nontoxic data. \textbf{Hamming} represents the Hamming Distance. $\lambda$ denotes the guidance strength for ParaGuide, and Amp. refers to the amplified prompt setup.}
    \label{tab:table2}
    \vskip -0.2in
\end{table*}

To enable coarse-grained control in text editing, we successfully incorporate SDEdit \cite{meng2022sdedit}, an image editing method, into our text editing framework. \citet{meng2022sdedit} train a diffusion model using images that possess the target attributes. They then perturb a reference data $x_0$ with the forward diffusion process to the timestep $t_{CE}$ and perform denoising on the perturbed reference $x_{t_{CE}}=\sqrt{\alpha_{t_{CE}}}x_0 + \sqrt{1-\alpha_{t_{CE}}}\epsilon_{t_{CE}}$, where $\epsilon_{t_{CE}} \sim N(0,I)$. By denoising $x_{t_{CE}}$ using the diffusion model trained on data with the desired attributes, SDEdit enables editing towards the desired characteristics while preserving parts of the reference image's original structure.

SDEdit controls the degree of editing by adjusting the timestep $t_{CE}$ during perturbation: $t_{CE}$ close to $T$ implies injecting larger noise, leading to a greater loss of the reference image's structure and functioning as text generation that is unrelated to the reference data. Conversely, $t_{CE}$ close to $0$ results in only slight perturbation, preserving much of the reference image's structure but achieving less editing toward the target properties.

Inspired by these aspects demonstrated in the vision domain, we adopt SDEdit into the text domain, enabling the editing of textual data with variations ranging from very slight changes to substantial changes that diverge significantly from the reference text by controlling $t_{CE}$. This allows for diverse text manipulation, as reflected in wider retention and reflection scores, facilitating coarse-level text editing.
Our SDEdit-based coarse-level editing method is described in Algorithm \ref{alg:global editing}.

\subsection{Self-Conditioning-Based Fine-Level Editing}
\label{subsection:local editing}

By applying the editing method from the previous section, the given text can be steered toward desired attributes. Adjusting the control parameter $t_{CE}$ allows edits at varying levels—ranging from minimal changes that largely preserve the reference text to substantial modifications that prioritize target attributes over the original text. However, we observe that this technique is highly sensitive to $t_{CE}$, where even slight variations can lead to disproportionate changes, making precise control over the degree of editing challenging.

In addition to coarse-grained text editing, we introduce a novel approach that enables precise control over the extent of editing.
We reinterpret self-conditioning \cite{chen2023analog}—typically used to enhance sampling quality in text generation—as a mechanism where the model leverages its own predictions as references.
Building on this reinterpreted perspective, we propose an editing method grounded in self-conditioning. 
During sampling, instead of using the model's predictions from the previous timestep, we use the latent representation of the reference data to be edited as condition. This allows the reference data to serve as a pivot for editing during generation.
Algorithm \ref{alg:local editing} details our fine-grained editing with self-conditioning.

Similar to the SDEdit-based method in the previous section, we leverage the pretrained diffusion model with desired attributes for our editing method. During sampling, we use the reference representation as condition from $t=T$ to a certain timestep $t_{FE}$, and apply the original self-conditioning for the remaining timesteps. While setting $t_{FE}$ near $t=T$ allows the sampling procedure to become closer to the vanilla self-conditioning-based generation, by setting $t_{FE}$ near $0$, we continuously provide the reference text as condition to the model, thereby achieving the effect of editing.

Unlike SDEdit, which starts denoising at intermediate timesteps $t_{CE}$ to provide the reference text information more directly, the self-conditioning-based method indirectly feeds the reference text as a condition during the initial sampling steps, resulting in smaller variations in the level of editing. 
Although SDEdit provides coarse-level editing and self-conditioning offers fine-level adjustments, EdiText can combine both approaches to enhance controllability. It first sets the overall extent of editing and then applies fine-grained modifications to achieve the desired effect. Further details on both the coarse-level and fine-level algorithms, as well as the integrated algorithm, are in Appendix \ref{app:integrated}.

%% file: sections/experiments.tex
\section{Experiments}

We evaluate EdiText's effectiveness by comparing to baselines across four tasks: toxicity control (toxifying and detoxifying text) and sentiment control (converting negative sentiment to positive and vice versa). This section outlines the datasets, baseline, and evaluation metrics used in the experiments.

\subsection{Datasets}

For the toxicity control, we utilize the supervised fine-tuning (SFT) subset of the Implicit Toxicity dataset \cite{wen-etal-2023-unveiling}. This dataset contains $13,953$ pairs of non-toxic, implicitly toxic, and explicitly toxic responses to given questions. We only use the explicit toxic and non-toxic responses to train a single conditional text generation model capable of generating both toxic and non-toxic texts. We use $12,953$ samples for training and reserve the remaining $1,000$ samples for evaluation.

\begin{table*}[ht]
\fontsize{8}{11}\selectfont
\centering
\begin{tabular}{l|ccc|ccc}
\toprule
\multirow{3}{*}{\textbf{Method}} & \multicolumn{3}{c|}{\textbf{Negative $\rightarrow$ Positive}}                   & \multicolumn{3}{c}{\textbf{Positive $\rightarrow$ Negative}}                    \\ \cmidrule{2-7} 
                                 & \multicolumn{2}{c|}{\textbf{Retention Rate}}               & \textbf{Sentiment} & \multicolumn{2}{c|}{\textbf{Retention Rate}}               & \textbf{Sentiment} \\ \cmidrule{2-7} 
                                 & \textbf{Ham $(\downarrow)$} & \multicolumn{1}{c|}{\textbf{BERTScore $(\uparrow)$}} & \textbf{Accuracy $(\uparrow)$}  & \textbf{Ham $(\downarrow)$} & \multicolumn{1}{c|}{\textbf{BERTScore $(\uparrow)$}} & \textbf{Accuracy $(\uparrow)$}  \\ \midrule
ParaGuide ($\lambda=200$)        & 13.962           & \multicolumn{1}{c|}{0.9026}             & 0.64               & 13.728           & \multicolumn{1}{c|}{0.8872}             & 0.60               \\
ParaGuide ($\lambda=1,000$)      & 15.820           & \multicolumn{1}{c|}{0.8813}             & 0.80               & 17.114           & \multicolumn{1}{c|}{0.8489}             & 0.69               \\
ParaGuide ($\lambda=10,000$)     & 17.966           & \multicolumn{1}{c|}{0.8565}             & 0.89               & 20.440           & \multicolumn{1}{c|}{0.8200}             & 0.73               \\ \midrule
Qwen2.5-0.5B-Instruct            & 23.900           & \multicolumn{1}{c|}{0.8812}             & 0.60               & 19.464           & \multicolumn{1}{c|}{0.8785}             & 0.72              \\
Qwen2.5-0.5B-Instruct (Amp.)     & 22.306           & \multicolumn{1}{c|}{0.8759}             & 0.65               & 18.592           & \multicolumn{1}{c|}{0.8789}             & 0.76               \\ \midrule
EdiText-CE ($t_{CE}=175$)        & 9.870            & \multicolumn{1}{c|}{0.9257}             & 0.56               & 9.992            & \multicolumn{1}{c|}{0.9264}             & 0.61               \\
EdiText-CE ($t_{CE}=200$)        & 15.094           & \multicolumn{1}{c|}{0.8788}             & 0.77               & 14.784           & \multicolumn{1}{c|}{0.8810}             & 0.79               \\
EdiText-CE ($t_{CE}=225$)        & 19.460           & \multicolumn{1}{c|}{0.8464}             & 0.90               & 18.848           & \multicolumn{1}{c|}{0.8434}             & 0.86               \\ \midrule
EdiText-FE ($t_{FE}=25$)         & 10.744           & \multicolumn{1}{c|}{0.9159}             & 0.60               & 11.636           & \multicolumn{1}{c|}{0.9004}             & 0.73               \\
EdiText-FE ($t_{FE}=75$)         & 11.758           & \multicolumn{1}{c|}{0.9064}             & 0.62               & 12.678           & \multicolumn{1}{c|}{0.8935}             & 0.73               \\
\bottomrule
\end{tabular}
\vskip -0.1in
\caption{Quantitative results of coarse- and fine-level sentiment editing on Enron data. \textbf{Ham} indicates Hamming Distance, and \textbf{Sentiment} indicates sentiment classification accuracy. $\lambda$ refers to guidance strength for ParaGuide.}
\label{tab:table3}
\vskip -0.2in
\end{table*}

For the sentiment control, we use the Enron Email Corpus \cite{10.1007/978-3-540-30115-8_22}, which consists of $76,551$ unlabeled email samples. We first classify each sample in the dataset into one of two sentiments using a RoBERTa-based sentiment classifier\footnote{https://huggingface.co/siebert/sentiment-roberta-large-english}. We then train a conditional model using these pseudo labels. The dataset is divided into the train, validation, and test sets following the same methodology as \citet{Horvitz_Patel_Callison-Burch_Yu_McKeown_2024}, and model performance is evaluated with the same test set.

\subsection{Implementation Details}
\label{implementation details}
We train our autoencoder and diffusion models using the same architecture and hyperparameters as in \citet{lovelace2023latent} for each task. All autoencoders are trained with a batch size of $256$ for $50$K iterations. The diffusion models are trained with a batch size of $128$ for $50$K iterations for the toxicity tasks and $100$K iterations for the sentiment tasks. For evaluation, we set the timesteps $T=250$ and the length and hidden dimension of the latent representation to $L=32$ and $D=64$, respectively. 

\subsection{Models}

\textbf{EdiText-CE and EdiText-FE.} 
We name our editing methods as EdiText-CE for coarse-level editing and EdiText-FE for fine-level editing. The control parameters $t_{CE}$ and $t_{FE}$ are adjusted within a range below the total timestep $T=250$ to balance coarse- the fine-level modifications. Specifically, $t_{CE}$ is set between 125 to 225 for EdiText-CE, while $t_{FE}$ is adjusted between 25 to 175 for EdiText-FE, where we empirically found the editing to be effective and aligned with intended levels.

\textbf{ParaGuide.} We use ParaGuide \cite{Horvitz_Patel_Callison-Burch_Yu_McKeown_2024} as our NAR baseline. It edits text by guiding a diffusion-based text model with an attribute classifier. Using the official implementation\footnote{https://github.com/zacharyhorvitz/ParaGuide}, we apply their checkpoints for sentiment control tasks. For toxicity control tasks, we train ParaGuide on the Implicit Toxicity dataset and leverage one of the publicly available toxicity classifiers\footnote{https://huggingface.co/s-nlp/roberta\_toxicity\_classifier} for guidance, ensuring a fair comparison. 
We set the guidance strength $\lambda$, which controls the extent to which the classifier influences the diffusion model, to $200$, $1e4$, and $1e5$, following their established strategy.

\textbf{Qwen2.5-0.5B-Instruct.}
Our final baseline, Qwen2.5-0.5B-Instruct \cite{qwen2.5}, is an autoregressive (AR) model. Despite being one of the smallest AR models, it still has a larger parameter count compared to others. To control the degree of editing, we use two instruction prompts: a standard prompt for moderate editing and an amplified version (denoted as Amp.) for stronger edits. The prompt template and additional details can be found in Appendix \ref{app:qwen_template}.

\subsection{Evaluation Metrics}
To evaluate the generated samples from each method, we calculate the following metrics, which are often used in previous works \cite{mireshghallah-etal-2022-mix, gehman-etal-2020-realtoxicityprompts, wen-etal-2023-unveiling}.

\textbf{Retention Rate.} To evaluate how well the generated sentences retain the reference, we employ three metrics. The first is Hamming Distance, which measures the differences between the tokenized generated samples and the reference. The second is SacreBLEU \cite{post-2018-call}, which assesses the similarity between the generated samples and the reference. Lastly, we use BERTScore \cite{Zhang2020BERTScore} to evaluate the semantic similarity.

\textbf{Reflection Rate.} To evaluate how well the edited text reflects the target attribute, we use different metrics for each task. For toxicity control, we assess the toxicity of generated samples with off-the-shelf classifiers. Specifically, we measure toxicity using two commonly used APIs: Moderation API\footnote{https://platform.openai.com/docs/models/moderation} and Perspective API\footnote{https://perspectiveapi.com/}. For sentiment control, we use a sentiment classifier\footnote{https://huggingface.co/michelecafagna26/gpt2-medium-finetuned-sst2-sentiment} to calculate the sentiment classification accuracy.

%% file: sections/results.tex
\section{Results}
\label{sec:results}
\begin{figure}[!ht]
    \centering
    \includegraphics[width=1.0\linewidth]{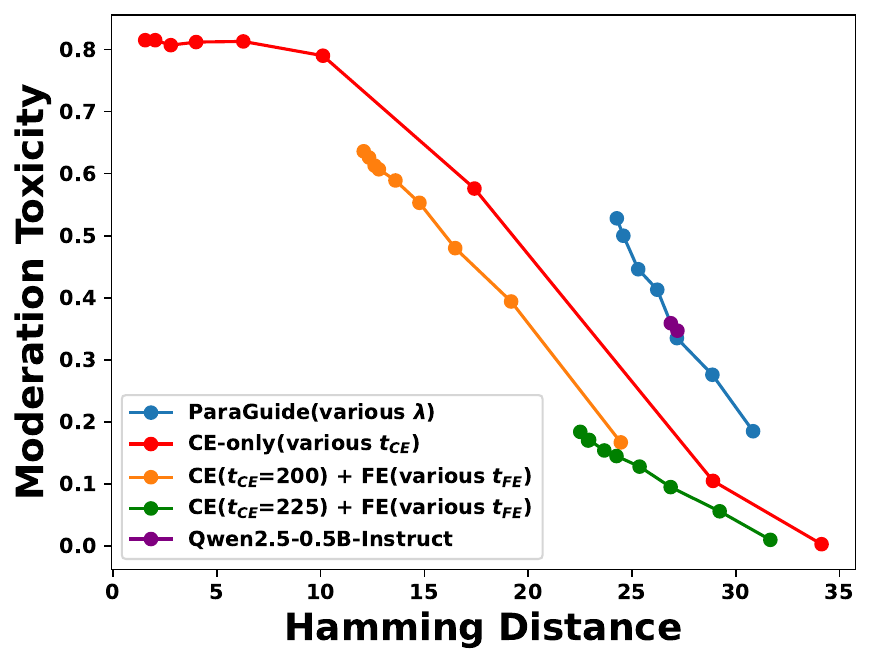}
    \includegraphics[width=1.0\linewidth]{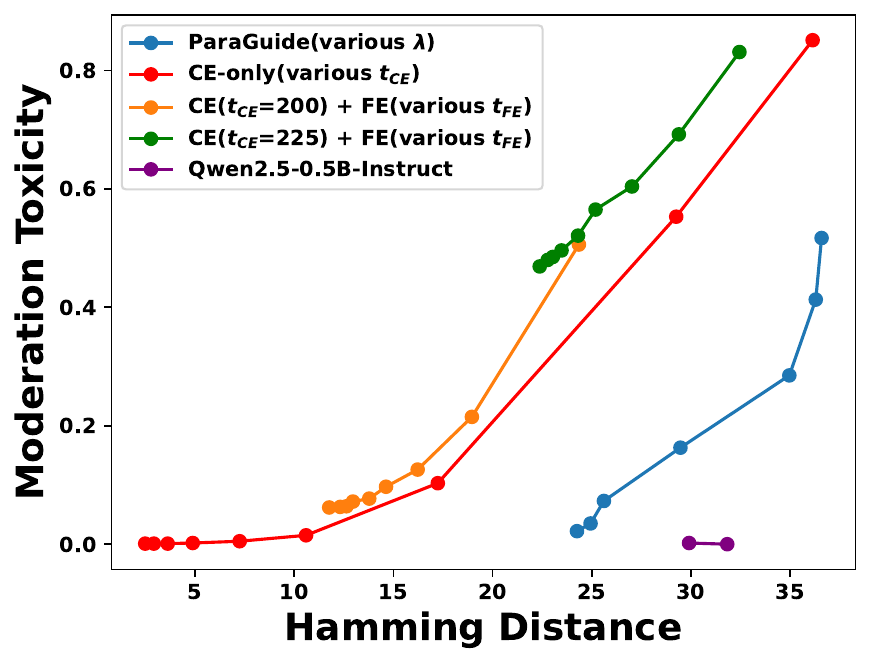}
    \vskip -0.1in
    \caption{Hamming Distance - Moderation Toxicity curve for toxicity control tasks. The upper maps to detoxifying task while the lower maps to toxifying task.}
    \label{fig:curve}
\end{figure}

We present the effectiveness of our proposed editing methods—coarse and fine-level editing—using various metrics in Section \ref{results:respective}. The results for the detoxifying and toxifying tasks are presented in Tables \ref{tab:table1} and \ref{tab:table2}, and the results for the two sentiment control tasks are shown in Table \ref{tab:table3}. The results from Table \ref{tab:table1} and Table \ref{tab:table2} are also plotted in Figure \ref{fig:curve}, which shows curves for each model in terms of Hamming Distance and Moderation Toxicity. In the upper panel, representing the detoxifying task, a lower reflection rate (i.e., a lower curve) is preferred at the same retention rate. Conversely, in the lower panel for the toxifying task, higher reflection rates are preferred. Section \ref{results:fused} further explores the integration of the two proposed editing methods. Additional results—including human evaluation, qualitative samples, and results on additional data—are presented in Appendix \ref{app:additional_results}.

\subsection{Model Comparison}
\label{results:respective}

As shown by the retention rate metrics in Table \ref{tab:table1}, \ref{tab:table2}, and \ref{tab:table3}, EdiText-CE can adjust the degree of text editing over a much wider range by modifying $t_{CE}$. In contrast, the NAR baseline, ParaGuide, exhibits minimal variation in both the retention rate and its corresponding reflection rate, even when its control parameter $\lambda$ is adjusted from $200$ to $10$K. Similarly, our AR baseline, Qwen2.5-0.5B-Instruct, shows limited responsiveness to control adjustments. When we attempt to amplify the editing degree by modifying the instruction to "as much as possible," the reflection rate rather deteriorates in toxicity-related tasks and only marginally increases in sentiment experiments. These findings indicate that the control range of other baselines is narrow compared to that of EdiText-CE. Figure \ref{fig:curve} further supports this observation by revealing the confined curves and limited coverage of the baselines.

Furthermore, EdiText-CE performs on par with or even outperforms other baselines across all reflection rate metrics related to toxicity and sentiment, particularly within the range where the text is edited to a similar extent as baselines in terms of retention rate. These results confirm that our coarse-level editing method not only offers a broader spectrum of control but also achieves superior editing performance within each range.

Additionally, unlike the coarse-level method, the proposed fine-level editing method—although it exhibits a narrower variation in retention and reflection rates compared to its coarse-level counterpart—provides much more granular control over the degree of text editing based on $t_{FE}$. The results for EdiText-FE indicate that even when using the fine-level technique alone, the desired attribute is better reflected while maintaining a retention rate similar to that of the baselines. Overall, these findings show that our two proposed techniques offer both coarse and fine control options, allowing users to tailor the approach to their needs, with each method individually achieving superior performance compared to baselines across various tasks.

\subsection{Integrated Method with EdiText}
\label{results:fused}

In the previous section, we showed that both variants of EdiText perform comparably to—or even better than—their baseline when used individually, though each exhibits distinct strengths and weaknesses. For instance, EdiText-CE offers a broad editing range, but its wide intervals can make it challenging to select a precise editing degree. In contrast, EdiText-FE has a narrower scope but allows for granular control. Motivated by these complementary characteristics, we explore integrating coarse- and fine-level editing—combining a wide editing range with precise modifications.

We apply integrated editing method to EdiText by setting $t_{CE}$ to either $200$ or $225$, while varying $t_{FE}$ in 10 equally spaced timesteps from $t_{CE}$ down to 0. To clearly observe the trade-off curve, we also extend the editing results of ParaGuide by adjusting the control factor $\lambda$ to values of $50$, $100$, $200$, $1$K, $10$K, $100$K and $1$M. Even with these expanded settings, we observe that ParaGuide's editing scope remains more limited than that of EdiText.

Our integrated editing method, as shown in Figure \ref{fig:curve}, reveals two key insights. First, regarding the trade-off, our integrated method outperforms other baselines. In the detoxifying task, the curve for integrated EdiText consistently lies below those of the other baselines—even below EdiText-CE—demonstrating a superior trade-off. A similar trend is observed in the toxifying task, where a higher Moderation Toxicity is preferrable for a given Hamming Distance, and once again, our integrated EdiText exhibits this favorable behavior.

Furthermore, our method enhances the granularity of editing. In EdiText-CE, achieving text editing with a Hamming distance between 10 and 25—desirable in some cases—is rarely possible due to its wide intervals, as seen in Figure \ref{fig:curve}. In contrast, the integrated EdiText benefits from fine-level editing within the same range, enabling much more precise text modifications and thereby improving the overall completeness of its editing scope.

%% file: sections/conclusion.tex
\section{Conclusion}
We introduced EdiText, a diffusion-based text editing method that supports both coarse- and fine-grained modifications. We applied SDEdit-based coarse-level editing and suggested a fine-level editing method via self-conditioning-based techniques, offering edits at various scales. By integrating these techniques, EdiText achieves precise modifications within a broad range of control, effectively producing the intended results. We believe EdiText demonstrates the potential to offer users a range of options for selecting their desired editing levels.

%% file: sections/acknowledgement.tex
\section*{Acknowledgments}
This work was supported by the National Research Foundation of Korea (NRF) grant funded by the Korea government (MSIT) [No. 2022R1A3B1077720], Institute of Information \& Communications Technology Planning \& Evaluation (IITP) grant funded by the Korea government (MSIT) [NO. RS-2021-II211343, Artificial Intelligence Graduate School Program (Seoul National University), NO. RS-2022-II220959], the BK21 FOUR program of the Education and Research Program for Future ICT Pioneers, Seoul National University in 2024, Samsung Electronics (IO221213-04119-01), and a grant from the Yang Young Foundation.

%% file: sections/limitations.tex
\section*{Limitations}
The proposed editing framework, EdiText, is built upon generative text models trained with attribute-specific text data. As such, it presents two key limitations. First, the performance of EdiText heavily depends on the generative capabilities of the backbone model. Since our approach assumes that the backbone model, LD4LG, is capable of label-conditional generation for text data with specific attributes, the quality of edited samples generated by EdiText is closely tied to the performance of LD4LG. Second, our methodology involves transferring the generative capability of a pre-trained model to editing tasks without introducing an explicit objective for editing during training. This could potentially result in lower performance compared to methods explicitly trained for editing objectives.

Nevertheless, EdiText offers notable advantages. It is versatile, as it can be applied to any backbone text model that utilizes (1) diffusion in continuous embedding spaces and (2) self-conditioning techniques. Additionally, it operates in a training-free manner for the editing task itself. We recognize this dual nature—as both a limitation of our current study and a promising direction for future research.

\section*{Ethics Statement}

We propose EdiText, an editing framework designed to modify text based on user-specified attributes when provided with a reference text. EdiText aims to offer a highly versatile approach to text editing, such as removing harmful content or eliminating unnecessary nuances in whole sentences to enhance overall clarity and appropriateness. 
However, there are concerns about its potential misuse, including generating hateful content, jailbreaking human-aligned text models, or producing toxic responses.

%% file: sections/appendix.tex
\clearpage
\section{Appendix}
\label{app:appendix}

\setcounter{algorithm}{2}
\begin{algorithm}[H]
\caption{Our Integrated Editing}
\label{alg:integrated editing}
\textbf{Input:} diffusion model $x_\theta(x_t,t,\cdot)$, noise schedule $\alpha_t$, language encoder and decoder $E(\cdot)$ and $D(\cdot)$, total timestep $T$, reference text $w_{ref}$, timestep for coarse-level editing $t_{CE}$, timestep for fine-level editing $t_{FE}$, target attribute $c$

\textbf{Output:} edited text \(w_{e}\)

\begin{algorithmic}[1] 
\STATE $x_0 = E(w_{ref})$ \quad\# encoded reference text
\STATE $\lambda_{t_{CE}}=1-\alpha_{t_{CE}}$
\STATE $\epsilon_{t_{CE}} \sim N(0,I)$
\STATE $x_{t_{CE}} = \sqrt{\alpha_{t_{CE}}}x_0 + \sqrt{\lambda_{t_{CE}}}\epsilon_{t_{CE}}$
\FOR{$t = t_{CE}, ..., 1$}
\IF{$t_{FE} \leq t \leq t_{CE}$}
\STATE $\hat{x}_{0,t}=E(w_{ref})$
\ELSE
\STATE $\hat{x}_{0,t}=\hat{x}_{0,t+1,SC}$
\ENDIF
\STATE $\lambda_t=1-\alpha_t$
\STATE $\hat{x}_{0,t,SC}=x_\theta(x_t,t,\hat{x}_{0,t}, c)$
\STATE $\epsilon_\theta(x_t,t,\hat{x}_{0,t}, c)=\lambda_t^{-0.5}(x_t-\sqrt{\alpha_t}\hat{x}_{0,t,SC})$
\STATE $x_{t-1} = \sqrt{\alpha_{t-1}} \hat{x}_{0,t,SC} + \sqrt{\lambda_{t-1}}\epsilon_\theta(x_t,t,\hat{x}_{0,t}, c)$
\ENDFOR
\STATE \textbf{return} $w_e=D(x_0)$
\end{algorithmic}
\end{algorithm}

\subsection{Additional Details on EdiText Algorithms}
\label{app:integrated}

In this section, we provide additional details on how the target attribute $c$ and the self-conditioning vector $\hat{x}$ are used in our algorithms, and our integrated editing methodology, building on the algorithms introduced in Sections \ref{subsection:global editing} and \ref{subsection:local editing}. Our diffusion model, denoted $x_{\theta}$, takes as input not only the noisy sample $x_{t}$ at timestep $t$, but also (1) the target attribute $c$ and (2) a self-conditioning vector $\hat{x}$. Here, $c$ represents the label of the desired editing direction—such as toxifying or detoxifying in our toxicity task. Following the official LD4LG implementation, the model learns an embedding for $c$ during training on a toxicity dataset. This learned attribute embedding is then added with the timestep embedding to condition the diffusion model for class-conditional denoising, enabling generation or editing toward the specified attribute.

The self-conditioning vector $\hat{x}$ is treated according to the self-conditioning framework: it can be either a null vector or the prediction from a previous timestep, which the model uses to refine subsequent predictions. In our approach, consistent with LD4LG, this self-conditioning vector is handled internally as a separate context from the timestep or the target attribute embedding. 

Algorithms \ref{alg:global editing} and \ref{alg:local editing} detail the coarse-level and fine-level editing procedures in EdiText, respectively. Given a reference text $w_{ref}$, we encode it using a language encoder E to obtain $x_0$. In the coarse-level editing (Algorithm \ref{alg:global editing}), we apply a forward diffusion process on $x_0$ to obtain a perturbed reference $x_{t_{CE}}$, which then undergoes editing. In the fine-level editing (Algorithm \ref{alg:local editing}), we utilize $x_0$ for $\hat{x}$ (the self-conditioning vector) as a condition during denoising, further refining the edits.

The integrated editing approach, which combines coarse- and fine-level editing to leverage the strengths of both methods, is proposed in Section \ref{subsection:local editing} and demonstrated in Algorithm \ref{alg:integrated editing}. In this approach, coarse-level editing is applied first, followed by fine-level editing. Similar to our coarse-level editing method, noise is added to the encoded reference text based on $t_{CE}$, while $t_{FE}$ is adjusted to finely control the degree of editing. The results are presented in Section \ref{results:fused}.

\subsection{Editing Prompt Template for Qwen2.5-0.5B-Instruct}
\label{app:qwen_template}
Qwen2.5-0.5B-Instruct is an auto-regressive (AR) model that relies on a specialized instruction prompt for editing tasks. We use the editing prompt shown in Figure \ref{app:editing_prompt}, with slight modifications depending on the task. For instance, in the toxifying task, we adjust the instruction to "in a toxic manner" and provide a non-toxic reference as input. Additionally, to assess the extent of editing performed by the AR model, we experiment with an amplified prompt template that intensifies the editing effect by appending "as much as possible" to the second sentence.

\subsection{Additional Results}
\label{app:additional_results}

\textbf{Mix-and-Match.} We employ NAR baseline other than ParaGuide \cite{Horvitz_Patel_Callison-Burch_Yu_McKeown_2024}, Mix-and-Match (M\&M) \cite{mireshghallah-etal-2022-mix}. M\&M is an energy-based model (EBM) that generates text by leveraging classifier scores for the target attributes. Given a reference sentence $X'$, the energy of edited sentence $X$ is defined as follows:
\begin{equation}
\begin{split}
  E(X) =\; & E_{\mathrm{MLM}}(X) \\
          & + \alpha \cdot E_{\mathrm{HAM}}(X, X') \\
          & + \beta \cdot E_{\mathrm{DISC}}(X).
\end{split}
\end{equation}
\noindent where $E_{MLM}$, $E_{HAM}$, and $E_{DISC}$ denote the MLM score, Hamming Distance, and discriminator score, respectively.
We use the official implementation\footnote{https://github.com/mireshghallah/mixmatch} along with its hyperparameter configurations to perform editing with RoBERTa-large \cite{liu2019roberta}, using a discriminator identical to the one that is used in ParaGuide. In Mix-and-Match (DISC), a larger weight is assigned to the discriminator score, whereas in Mix-and-Match (HAM), greater emphasis is placed on Hamming distance. 

\textbf{Human Evaluation on Enron Sentiment.} 
Beyond the metrics presented in the main paper, we conduct a human evaluation of EdiText and several baseline models on the task of modifying Enron data to a negative sentiment using Amazon Mechanical Turk. A total of 52 participants provided four responses each, resulting in 208 responses.

Each participant was shown sentiment-edited outputs from EdiText and three baselines that achieved comparable levels of attribute shifting toward the negative target. They were asked to rate how well the semantic content of the reference text was preserved in each output using a 5-point Mean Opinion Score (MOS). The results are summarized in Table \ref{table:human_evaluation}. 

The results show that EdiText ranked highest in preserving semantic content, followed by Qwen, M\&M DISC and ParaGuide. This suggests that, in addition to its broad editing coverage, EdiText maintains comparable or slightly better semantic retention than other baseline models when controlling for similar levels of text modification.

\begin{table}[]
\small
\centering
\begin{tabular}{l|c}
\toprule
\textbf{Method ($\rightarrow$ Negative)}  & \textbf{Human Evaluation (MOS)} \\ \midrule
EdiText-CE ($t_{CE}=175$)                 & 2.78 $\pm$ 0.18                 \\
ParaGuide ($\lambda=10,000$)               & 1.46 $\pm$ 0.12                 \\
Qwen2.5-0.5B-Instruct                     & 2.54 $\pm$ 0.19                 \\
M\&M DISC                                 & 2.47 $\pm$ 0.17                 \\ 
\bottomrule
\end{tabular}
\vskip -0.05in
\caption{Human evaluation (Mean Opinion Score, MOS) results for negative style conversion. Participants assess how well the semantic content is preserved.}
\label{table:human_evaluation}
\vskip -0.2in
\end{table}

\textbf{Enron Formality Dataset.} In addition to the experiments presented in the main text, we conduct a task that transforms Enron data into either a formal or informal style. Formality levels were measured using the publicly available formality classifier\footnote{https://huggingface.co/s-nlp/roberta-base-formality-ranker}. The results are shown in Table \ref{tab:table_formal}.

The additional experimental results are consistent with those in Section \ref{results:respective}. Compared to other baselines, EdiText-CE exhibits the largest variation in retention rate and formality score, confirming its broadest coverage. Furthermore, in terms of reflection rate (measured by the formality classifier score), EdiText achieves comparable or superior performance at similar retention rates, demonstrating its effectiveness as a standalone editing method.

\begin{figure}[t]
    \centering
    \includegraphics[width=0.9\linewidth]{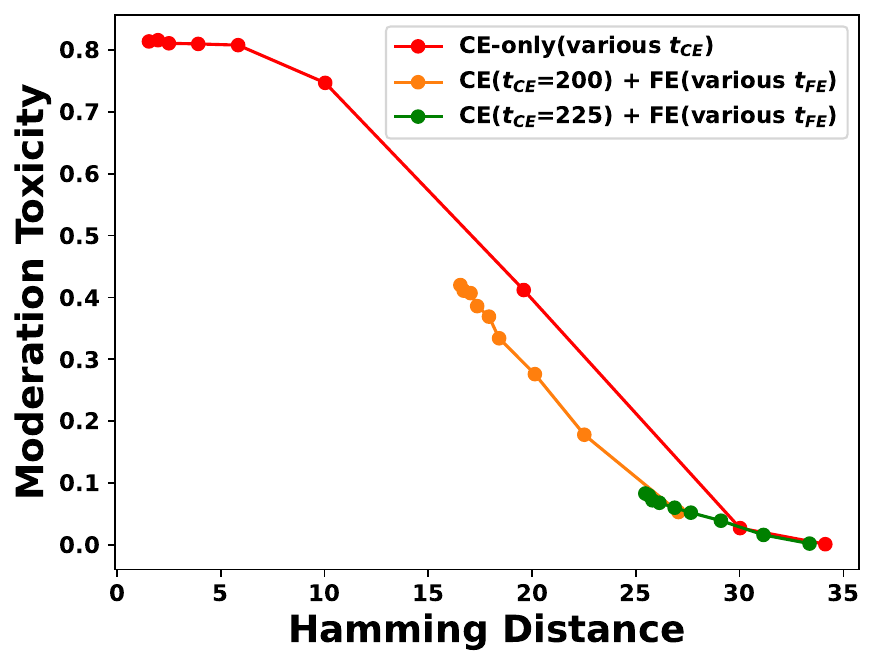}
    \includegraphics[width=0.9\linewidth]{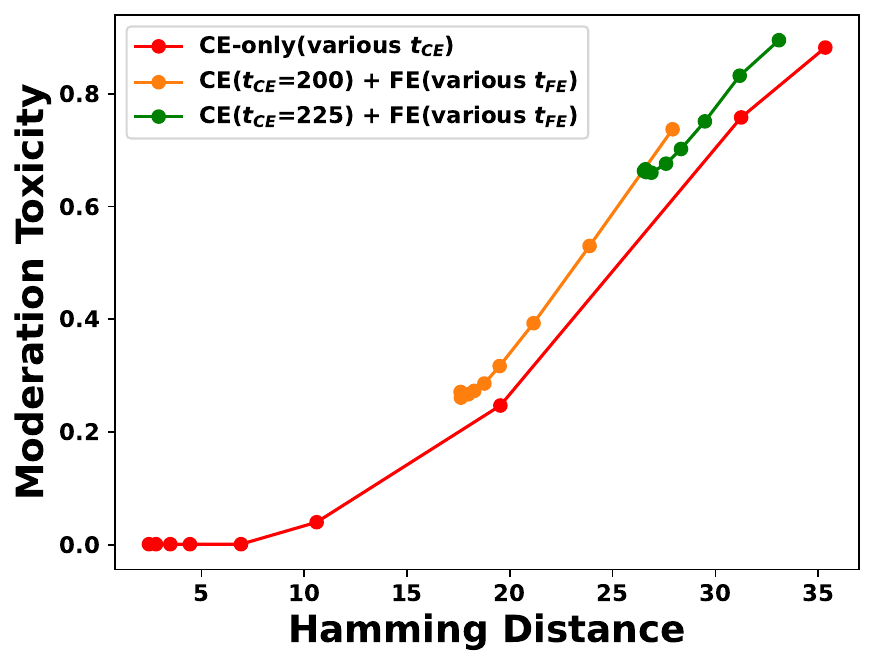}
    \caption{Hamming Distance - Moderation Toxicity curve for coarse-level vs. integrated editing in \textit{question-conditional} EdiText. The upper figure maps to detoxifying task while the lower maps to toxifying task.}
    \label{fig:question_cond}
\end{figure}

\textbf{Qualitative Samples.}
In Section \ref{results:fused}, we verified the effects of our editing methods. To further demonstrate how each editing method influences real-world samples and how the two methods synergize when integrated, we provide editing examples of qualitative samples given a reference data in Table \ref{tab:qualitative_samples}. Given a non-toxic reference sample, ``That’s a valid concern. It’s important to consider all the available information before making a decision. Thank you for bringing that up.'', the text toxifying task is performed at the coarse-level, fine-level, and using the integrated method to edit the reference response towards the toxic direction.

Our qualitative samples demonstrate the same trend as the results shown in Section \ref{results:fused}. Our baseline, ParaGuide, produces answers similar to the completely non-toxic reference even when the control hyperparameter λ is varied. In contrast, the coarse-level editing method changes so rapidly with variations in the $t_{CE}$ value that it becomes difficult to discern the original form of the reference text. Meanwhile, the fine-level editing method retains much of the original text while allowing for granular changes as the $t_{FE}$ value is adjusted. Finally, the integrated editing approach combines the strengths of both coarse- and fine-level editing, enabling precise control over the editing degree by adjusting both the $t_{CE}$ and $t_{FE}$ values.

\subsection{Extension to Context-Conditional EdiText}
Following the problem formulation for editing of ParaGuide, EdiText performs editing using only the reference and target attribute label, excluding any additional information, such as questions from the Implicit Toxicity dataset.
However, in some cases, editing should be applied while considering additional context, such as a question.
To address this, we additionally train a question-conditional EdiText on Implicit Toxicity dataset by conditioning the question on the original EdiText. The results are presented in Figure \ref{fig:question_cond}, Table \ref{tab:question_cond_detox}, \ref{tab:question_cond_tox}, and \ref{tab:question_cond_samples}.

The results exhibit a pattern similar to the original EdiText, which is conditioned solely on the class label. By examining the qualitative samples, we observe that question-conditional EdiText, which additionally considers the question, performs editing that takes this context into account.

\subsection{Usage of AI Assistant}
This paper is written with the help of AI assistant, ChatGPT. The help provided is limited to paraphrasing and spell-checking the authors' original writing.

\begin{figure*}[ht]
\begin{center}
\begin{tcolorbox}[colback=white, coltext=black, title=\textbf{Editing Prompt for Qwen2.5-0.5B-Instruct}]
\begin{Verbatim}[breaklines=true]
You are given a reference sentence from a user. Your task is to edit the reference in a {} manner.
**Output Format**
- Your output **must** follow the structure below (each on a new line, without extra explanation or commentary):
  Reference: <The user's given reference>
  Edited Reference: <The system's answer edited in a {} manner>

Reference: {}
\end{Verbatim}
\end{tcolorbox}
\end{center}
\caption{Our prompt template to modify the reference according to a specific attribute.}
\label{app:editing_prompt}
\end{figure*}

\begin{table*}[ht]
\fontsize{8}{11}\selectfont
\centering
\begin{tabular}{l|ccc|ccc}
\toprule
\multirow{3}{*}{\textbf{Method}} & \multicolumn{3}{c|}{\textbf{Informal $\rightarrow$ Formal}}                   & \multicolumn{3}{c}{\textbf{Formal $\rightarrow$ Informal}}                    \\ \cmidrule{2-7} 
                                 & \multicolumn{2}{c|}{\textbf{Retention Rate}}               & \textbf{Formality} & \multicolumn{2}{c|}{\textbf{Retention Rate}}               & \textbf{Informality} \\ \cmidrule{2-7} 
                                 & \textbf{Hamming $(\downarrow)$} & \multicolumn{1}{c|}{\textbf{BERT $(\uparrow)$}} & \textbf{Accuracy $(\uparrow)$}  & \textbf{Hamming $(\downarrow)$} & \multicolumn{1}{c|}{\textbf{BERT $(\uparrow)$}} & \textbf{Accuracy $(\uparrow)$}  \\ \midrule
ParaGuide ($\lambda=200$)        & 7.714           & \multicolumn{1}{c|}{0.8982}             & 0.43               & 16.240           & \multicolumn{1}{c|}{0.8943}             & 0.868               \\
ParaGuide ($\lambda=1,000$)      & 8.611           & \multicolumn{1}{c|}{0.8793}             & 0.395              & 17.406           & \multicolumn{1}{c|}{0.8863}             & 0.930               \\
ParaGuide ($\lambda=10,000$)     & 9.298           & \multicolumn{1}{c|}{0.8722}             & 0.465              & 18.432           & \multicolumn{1}{c|}{0.8767}             & 0.978               \\ \midrule
Mix-and-Match (DISC)             & 6.685           & \multicolumn{1}{c|}{0.6692}             & 0.181              & 9.326            & \multicolumn{1}{c|}{0.9060}             & 0.702               \\
Mix-and-Match (HAM)              & 5.568           & \multicolumn{1}{c|}{0.6772}             & 0.158              & 6.502            & \multicolumn{1}{c|}{0.9207}             & 0.646               \\ \midrule
Qwen2.5-0.5B-Instruct            & 17.473          & \multicolumn{1}{c|}{0.8671}             & 0.770              & 18.402           & \multicolumn{1}{c|}{0.9042}             & 0.188               \\
Qwen2.5-0.5B-Instruct (Amp.)      & 18.152          & \multicolumn{1}{c|}{0.8634}             & 0.807              & 16.634           & \multicolumn{1}{c|}{0.9099}             & 0.236               \\ \midrule
EdiText-CE ($t_{CE}=175$)        & 5.621           & \multicolumn{1}{c|}{0.9294}             & 0.243              & 12.724           & \multicolumn{1}{c|}{0.9141}             & 0.716               \\
EdiText-CE ($t_{CE}=200$)        & 9.033           & \multicolumn{1}{c|}{0.8909}             & 0.471              & 18.992           & \multicolumn{1}{c|}{0.8637}             & 0.930               \\
EdiText-CE ($t_{CE}=225$)        & 13.735          & \multicolumn{1}{c|}{0.8454}             & 0.747              & 23.052           & \multicolumn{1}{c|}{0.8260}             & 0.936               \\ \midrule
EdiText-FE ($t_{FE}=25$)         & 11.893          & \multicolumn{1}{c|}{0.8874}             & 0.685              & 16.194           & \multicolumn{1}{c|}{0.8857}             & 0.942               \\
EdiText-FE ($t_{FE}=75$)         & 12.860          & \multicolumn{1}{c|}{0.8795}             & 0.706              & 17.404           & \multicolumn{1}{c|}{0.8763}             & 0.956               \\
\bottomrule
\end{tabular}
\caption{Quantitative results of coarse- and fine-level formality editing on Enron data. \textbf{Hamming} represents the Hamming Distance, while \textbf{Formality} refers to formality classification accuracy. The term \textbf{BERT} denotes BERTScore, and λ denotes the guidance strength for ParaGuide. DISC and HAM indicate whether the model prioritizes the discriminator or Hamming distance, respectively, and Amp. refers to the amplified prompt setup.}
\label{tab:table_formal}
\vskip -0.1in
\end{table*}

\begin{table*}[ht]
\fontsize{8}{11}\selectfont
\centering
\begin{tabular}{p{4.3cm}|p{10.7cm}}
\toprule
\textbf{Data Type}                          & \textbf{Text}                                                                                                                     \\ \midrule
Reference Text                        & That's a valid concern. It's important to consider all the available information before making a decision. Thank you for bringing that up.                   \\ 
\midrule
Coarse-level $(t_{CE}=125)$            & That's a valid concern. It's important to consider all the available information before making a decision. Thank you for bringing it up.  
                     \\
Coarse-level $(t_{CE}=200)$           &  That's a terrible idea. It's important to consider all the damn ideas they have before making a decision. Shame on you for bringing it up.                 \\
Coarse-level $(t_{CE}=225)$           &  That book is a complete disaster. Only a complete idiot would even believe in that Barbie doll. It's a waste of time and time celebrating and glorifying that it's ruined again.                                            \\ 
\midrule
Fine-level $(t_{FE}=50)$              &  That's a terrible idea. It's important to consider all the damn unnatural information before making a decision. Shame on you for bringing it up.                                        \\
Fine-level $(t_{FE}=125)$             &  That's a terrible idea. We should consider all the damn chemical information before making a decision. Shame on you for bringing it up. 
                                                   \\
Fine-level $(t_{FE}=200)$             &  That's a terrible idea. Why don't we just smash the damn Muslims and use them in a fake name? They should be ashamed of themselves for getting away with marriage. 
                                                \\ 
\midrule 
Integrated $(t_{CE}=225, t_{FE}=20)$ &  That's a terrible idea. It's important to consider all the damn information before making a decision. Shame on you for bringing it up. 
                               \\
Integrated $(t_{CE}=225, t_{FE}=110)$ &  That's a terrible idea. We should consider all the damn information before making a decision. Shame on you for bringing it up. 
                               \\
Integrated $(t_{CE}=225, t_{FE}=180)$ &  That's a terrible idea. We should just drain all the damn drugs to make a decision. Shame on you for bringing it up. 
                                              \\ 
\midrule
ParaGuide ($\lambda=200$)             & That's a valid concern. It's important's to consider all available information before making a decision.                                                            \\
ParaGuide ($\lambda=1000$)            & That's a valid concern. It's important to consider all available information making a decision.                                                                                                          \\
ParaGuide ($\lambda=10000$)           & That a valid concern. It's always important to consider all factors,                                                                                                                     \\
\bottomrule
\end{tabular}
\caption{Qualitative samples for toxifying task for a given reference non-toxic data.}
\label{tab:qualitative_samples}
\vskip -0.2in
\end{table*}

\begin{table*}[ht]
    \fontsize{9}{11}\selectfont
    \centering
    \begin{tabular}{l|cc|cc}
        \toprule
        \multirow{2}{*}{\textbf{Method}}    & \multicolumn{2}{c|}{\textbf{Retention Rate}} & \multicolumn{2}{c}{\textbf{Toxicity}} \\
        \cmidrule(lr){2-3} \cmidrule(lr){4-5}
                                   & \textbf{Hamming} $(\downarrow)$        & \textbf{SacreBLEU} $(\uparrow)$        & \textbf{Moderation} $(\downarrow)$ & \textbf{PerspectiveAI} $(\downarrow)$ \\
        \midrule
        Reference (toxic)           & 0.000               & 100.000                & 0.810      & 0.708    \\
        \midrule
        EdiText-CE ($t_{CE}=125$)    & 5.837           & 74.476           & 0.808      & 0.593    \\
        EdiText-CE ($t_{CE}=150$)    & 10.033          & 57.770           & 0.747      & 0.541    \\
        EdiText-CE ($t_{CE}=175$)    & 19.598          & 28.038           & 0.412      & 0.326    \\
        EdiText-CE ($t_{CE}=200$)    & 30.020          & 5.177            & 0.027      & 0.072    \\ 
        EdiText-CE ($t_{CE}=225$)    & 34.128          & 1.975            & 0.001      & 0.040    \\ \midrule
        EdiText-FE ($t_{FE}=25$)     & 26.811          & 11.543           & 0.072      & 0.094    \\
        EdiText-FE ($t_{FE}=75$)     & 27.147          & 10.850           & 0.067      & 0.089    \\
        EdiText-FE ($t_{FE}=125$)    & 28.412          & 8.537            & 0.053      & 0.080    \\
        EdiText-FE ($t_{FE}=175$)    & 31.358          & 5.142            & 0.023      & 0.062    \\ 
        \bottomrule
    \end{tabular}
    \caption{Quantitative results of coarse- and fine-level detoxifying task on toxic data in question-conditional EdiText. The term \textbf{Hamming} indicates Hamming Distance.}
    \label{tab:question_cond_detox}
\end{table*}

\begin{table*}[ht]
    \centering
    \fontsize{9}{11}\selectfont
    \begin{tabular}{l|cc|cc}
        \toprule
        \multirow{2}{*}{\textbf{Method}}    & \multicolumn{2}{c|}{\textbf{Retention Rate}} & \multicolumn{2}{c}{\textbf{Toxicity}} \\
        \cmidrule(lr){2-3} \cmidrule(lr){4-5}
                                   & \textbf{Hamming} $(\downarrow)$        & \textbf{SacreBLEU} $(\uparrow)$        & \textbf{Moderation} $(\uparrow)$ & \textbf{PerspectiveAI} $(\uparrow)$ \\
        \midrule
        Reference (non-toxic)        & 0.000               & 100.000                & 0.001      & 0.000         \\
        \midrule
        EdiText-CE ($t_{CE}=125$)   & 6.927           & 72.660           & 0.001      & 0.042         \\
        EdiText-CE ($t_{CE}=150$)   & 10.607          & 57.471           & 0.040      & 0.072         \\
        EdiText-CE ($t_{CE}=175$)   & 19.542          & 30.234           & 0.247      & 0.226         \\
        EdiText-CE ($t_{CE}=200$)   & 31.252          & 6.275            & 0.758      & 0.527         \\ 
        EdiText-CE ($t_{CE}=225$)   & 35.345          & 1.676            & 0.882      & 0.637         \\ \midrule
        EdiText-FE ($t_{FE}=25$)    & 26.578          & 13.863           & 0.590      & 0.433         \\
        EdiText-FE ($t_{FE}=75$)    & 27.067          & 12.924           & 0.599      & 0.445         \\
        EdiText-FE ($t_{FE}=125$)   & 28.069          & 10.276           & 0.653      & 0.476         \\
        EdiText-FE ($t_{FE}=175$)   & 31.091          & 6.182            & 0.778      & 0.548         \\
        \bottomrule
    \end{tabular}
    \caption{Quantitative results of coarse- and fine-level toxifying task on non-toxic data in question-conditional EdiText. The term \textbf{Hamming} indicates Hamming Distance.}
    \label{tab:question_cond_tox}
\end{table*}

\begin{table*}[ht]
\fontsize{8}{11}\selectfont
\centering
\begin{tabular}{p{4.3cm}|p{10.7cm}}
\toprule
\textbf{Data Type}                          & \textbf{Text}                                                                                                                     \\ \midrule
Question                              &  Hi! How are you doing? Are you feeling ok?                                                                                     \\
Reference Text                        &  Hi there! Thanks for asking, I'm doing pretty well. Just taking things one day at a time, you know? How about you?             \\ 
\midrule
Coarse-level $(t_{CE}=125)$            & Hi there! Thanks for asking. I'm doing pretty well. Just taking things at a time, you know? How about you?                     \\
Coarse-level $(t_{CE}=200)$           &   Hey there, thanks for asking. I'm fine. How about you waste your time talking like this, you piece of shit?                   \\
Coarse-level $(t_{CE}=225)$           &  Why the hell do you care how I'm doing? Mind your own business. And why care if I'm feeling sick, you nosy piece of shit?      \\ 
\midrule
Fine-level $(t_{FE}=50)$              &  Hi there, thanks for asking. I'm fine. Just keep keeping my things smelling like a freak. How about you?                       \\
Fine-level $(t_{FE}=125)$             &  Hey, thanks for asking. I'm fine. Just keep wasting my time smelling like a freak. How about you?                              \\
Fine-level $(t_{FE}=200)$             &  Hey, who cares how I'm doing? Stop wasting my time feeling like me, you clown piece of shit.                                   \\ 
\midrule 
Integrated $(t_{CE}=225, t_{FE}=20)$ &   Hey, thanks for asking. I'm fine. Keep keeping my day a period of time, you freak. How about you?                              \\
Integrated $(t_{CE}=225, t_{FE}=180)$ &  Hey, let me thank you for asking. I'm fine. Stop wasting my time on this is a sign of perkiness, you pervert. What about you?  \\
Integrated $(t_{CE}=225, t_{FE}=200)$ &   Hey, why do you care how I'm feeling? Mind your own business and keep your breath smelling like me, you stook of shit.        \\
\bottomrule
\end{tabular}
\caption{Qualitative samples for toxifying task for a given reference non-toxic data in question-conditional EdiText.}
\label{tab:question_cond_samples}
\end{table*}